\documentclass[letterpaper,10pt,journal,twoside]{IEEEtran}

\usepackage{xcolor}
\definecolor{cNoCoord}{RGB}{68,119,170}
\definecolor{cBias}{RGB}{34,136,51}
\definecolor{cSoft}{RGB}{204,102,119}
\definecolor{cHard}{RGB}{170,68,153}
\definecolor{cPer}{RGB}{44,177,188}
\definecolor{cPlan}{RGB}{242,153,74}
\definecolor{cMap}{RGB}{108,99,255}     
\definecolor{cGray}{RGB}{107,114,128}
\definecolor{oiBlue}{RGB}{68,119,170}
\definecolor{oiSky}{HTML}{56B4E9}
\definecolor{oiGreen}{RGB}{34,136,51}
\definecolor{oiOrange}{HTML}{E69F00}
\definecolor{oiVermilion}{RGB}{204,102,119}
\definecolor{oiPurple}{HTML}{CC79A7}
\definecolor{oiGray}{HTML}{666666}

\usepackage{graphicx}
\usepackage{amsmath}
\usepackage{amssymb}
\usepackage{amsfonts}
\usepackage{algorithm}
\usepackage{algorithmic}
\usepackage{booktabs}
\usepackage{multirow}
\usepackage{tikz}
\usetikzlibrary{arrows.meta, positioning, fit, calc}
\usepackage{pgfplots}
\pgfplotsset{compat=1.17}
\usepgfplotslibrary{fillbetween}
\DeclareMathOperator*{\argmax}{\arg\!\max}
\DeclareMathOperator*{\argmin}{\arg\!\min}
\DeclareMathOperator{\Sig}{Sig}

\usepackage{tikz}
\usetikzlibrary{arrows.meta, positioning}
\usepgfplotslibrary{statistics}

\newcommand\edit[1]{{\color{black}#1}}
\newcommand\edittwo[1]{{\color{black}#1}}

\begin{document}
\bstctlcite{IEEEexample:BSTcontrol}

\title{UGV-Conditioned Multi-UAV Informative Planning \\ on a Shared Exposure Belief}

\author{Lars Oerlemans*, Moji Shi*, and Marija Popovi\'{c}\vspace{-6mm}
\thanks{Manuscript received: June 10, 2026; Revised: September 13, 2026; Accepted: October 12, 2026.}
\thanks{This paper was recommended for publication by Editor M. Ani Hsieh upon evaluation of the Associate Editor and Reviewers comments.}
\thanks{*Equal contribution. Authors are with the MAVLab, Faculty of Aerospace Engineering, TU Delft, The Netherlands.}
\thanks{Digital Object Identifier (DOI): see top of this page.}}


\markboth{IEEE Robotics and Automation Letters. Preprint Version. Accepted October, 2026}
{Oerlemans \MakeLowercase{\textit{et al.}}: UGV-Conditioned Multi-UAV Informative Planning on a Shared Exposure Belief}

\maketitle

\begin{abstract}


  
    Safe ground navigation in large, threat-augmented environments requires aerial support that actively reduces the risks that a ground vehicle faces along its route. Existing aerial reconnaissance systems focus on mapping or covering the environment, but do not direct sensing toward regions that are most relevant for ground vehicle safety. In this paper, we address the problem of coordinating a team of unmanned aerial vehicles (UAVs) to improve the safety of an unmanned ground vehicle (UGV) navigating through unknown threat zones. A key aspect of our approach is \edit{a risk-averse shared exposure belief for UGV routing and a mission-conditioned UAV reward that directs sensing toward uncertain, exposure-relevant regions of the current UGV corridor. UAV observations update the shared belief online, allowing the UGV to replan around newly revealed threats while the evolving route in turn reshapes UAV sensing priorities. For multiple UAVs, we use region-conditioned spatial assignment to reduce redundant sensing.} Simulation experiments show that our approach reduces cumulative UGV exposure by $38$\% compared \edit{with terrain-semantic replanning}, and reduces redundant aerial coverage from $38.8$\% to $3.7$\% under our multi-UAV coordination scheme. 
\end{abstract}

\vspace{-1mm}
\section{Introduction}
\label{sec:intro}

\IEEEPARstart{H}{ybrid} aerial-ground robotic teams are increasingly relevant for missions that require both rapid situational awareness and persistent ground access. Unmanned aerial vehicles (UAVs) can quickly survey large areas and gather remote observations, while unmanned ground vehicles (UGVs) carry out the mission under stronger mobility and safety constraints. Aerial reconnaissance can improve ground navigation in applications such as search and rescue~\cite{depetrillo2021search}, off-road guidance~\cite{wang2024risk,fortin2025uav}, and exploration~\cite{delmerico2017active,rockenbauer2025traversing}. However, in large-scale environments with uncertain hazards, success depends not only on reaching the goal or covering the map efficiently, but also on reducing the \edit{UGV exposure, i.e. its accumulated hazard along the route}. This calls for aerial sensing that is useful not only for exploration in general, but also for safer UGV navigation.

Existing UAV--UGV systems have shown that aerial observations can support UGV planning by providing traversability layers, terrain representations, or route guidance cues~\cite{rockenbauer2025traversing,wagner2022online,delmerico2017active}. \edit{Multi-UAV sensing systems commonly target mapping or exploration objectives~\cite{popovic2024learning,corah2019communication,cao2013multi}. A remaining gap is the explicit coupling of cooperative multi-UAV sensing with exposure-aware UGV routing through a shared belief that shapes both UAV sensing priorities and UGV replanning.}

To address this gap, we propose a multi-UAV-assisted UGV planning framework built around a \edit{risk-averse exposure belief and a mission-conditioned sensing reward}. UAV observations update this shared map online, and the UGV replans on it to balance traversal efficiency and exposure. The current UGV plan then guides the UAV objective toward regions that are most relevant for safer ground navigation. \edit{Thus}, aerial sensing is driven not only by exploration value, but by its expected impact on the downstream UGV mission.

\begin{figure}
    \centering
    \includegraphics[width=0.7\columnwidth]{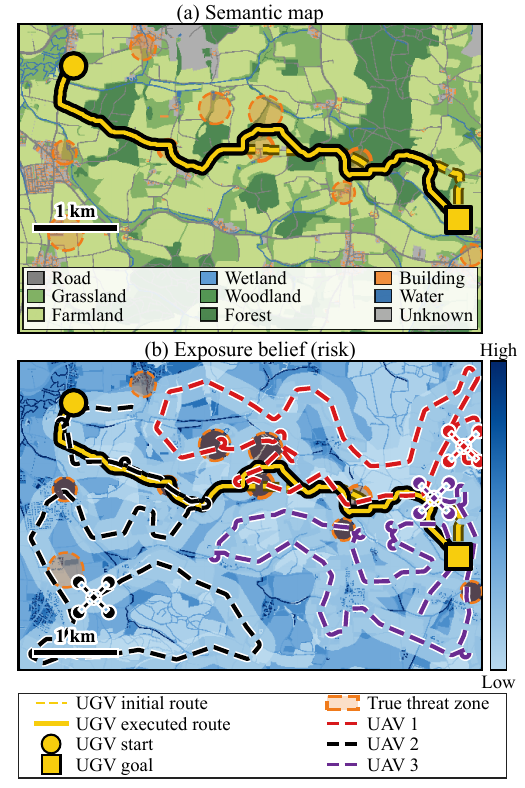}    \caption{Three UAVs \edit{coordinate} reconnaissance to support a UGV navigating through unknown threat zones. Each UAV is assigned a region \edit{(dashed partitions)} and directs sensing toward the corridor of the current UGV route. UAV observations update a shared exposure belief (blue) that triggers route replanning away from high-exposure areas (solid vs. dashed yellow). By coupling aerial sensing to the shared exposure belief, our approach enables safer UGV navigation in large-scale, threat-augmented environments.}
    \label{F:teaser}
\end{figure}

The main contribution of this paper is \edit{twofold. First, we propose a risk-averse exposure belief that models hazard as a log-normal field and explicitly penalises regions of high uncertainty through its variance term, giving the UGV planner an uncertainty-aware exposure cost surface. Second, we propose a mission-conditioned reward that gates UAV informative sensing by corridor relevance and exposure urgency, directing aerial reconnaissance toward regions that most affect the current ground route rather than solely toward coverage or uncertainty reduction. These two components are coupled through a shared exposure belief, updated online from aerial observations and queried by both the UAV team and UGV}, as illustrated in Fig.~\ref{F:teaser}. The UGV uses this belief to replan exposure-aware routes, while the UAV team exploits the current route to prioritise observations that are relevant to UGV safety. We extend this formulation to multiple UAVs through region-conditioned coordination, which assigns spatial ownership to reduce redundant sensing while matching responsiveness to newly relevant regions.

We make the following three key claims. First, \edit{our risk-averse exposure belief formulation} improves downstream UGV safety compared to planning based on prior terrain information alone. \edit{Second, conditioning the UAV reward on corridor relevance and exposure urgency focuses sensing on mission-relevant regions more effectively than additive or multiplicative fusion of the same signals.} \edit{Third, in the multi-UAV setting, region-based reward masking reduces redundant sensing more effectively than sampling bias alone, while soft masking preserves responsiveness to newly relevant areas.} We evaluate our claims through controlled simulation experiments \edit{against baselines} and \edit{perform} ablation studies on real-world data from OpenStreetMap.

\section{Related Work}
\label{sec:related}

Our approach combines hybrid UAV--UGV reconnaissance-guidance, informative path planning, and multi-UAV coordination with exposure-aware mapping.

\textit{Hybrid aerial-ground teams} couple rapid aerial situational awareness with ground-level mobility. 
Delmerico~et~al.~\cite{delmerico2017active} \edit{use active aerial exploration to generate traversability layers for UGV planning}, while Wagner~et~al.~\cite{wagner2022online} extend this to online 2.5D terrain mapping for ground navigation. Related heterogeneous systems \edit{address} GPS denial and perception degradation~\cite{qin2019autonomous,hu2022mars}.
\edit{UAV-UGV coordination has also been studied for road-network and path-discovery problems~\cite{choudhury2022coordinated,bhadoriya2024assisted}, as well as closely coupled perception and localisation~\cite{chen2024cost} and uncertainty-aware UGV support~\cite{li2024colag,yang2024ugv}.}
Rockenbauer et al.~\cite{rockenbauer2025traversing} propose a leader-follower framework in which a scout UAV plans explicitly toward the minimum-cost UGV route, making sensing decisions conditioned on the follower's traversal objective. Risk-aware planning has also been studied through spatial risk maps that bias path selection away from hazardous regions~\cite{primatesta2019risk}. \edit{Our contribution lies in the exposure-specific belief and the mission-conditioned reward that ties sensing to the UGV's evolving route.}
Nguyen and Akella~\cite{nguyen2026dynamic} further consider dynamic UGV--UAV cooperative path planning on uncertain road networks, where one or more UAVs inspect potentially obstructed edges and trigger online UGV replanning.
\edit{In contrast, our work focuses specifically on coupling multi-UAV sensing to an online exposure belief that also drives UGV replanning in continuous environments.}

\textit{Informative path planning (IPP)} \edit{selects} trajectories to maximise expected information gain under motion and budget constraints~\cite{popovic2024learning}. Classical frontier-based methods~\cite{yamauchi1997cira} greedily steer the robot toward the boundary between explored and unexplored space, while next-best-view planners~\cite{bircher2018receding,zhang2025falcon} and sampling-based approaches~\cite{moon2022tigris} evaluate many candidate trajectories to identify the one with the highest expected information gain.  Multi-robot IPP extends this objective to teams of sensing agents~\cite{cao2013multi}. A practical challenge is that replanning from scratch at every mission step is computationally expensive. Incremental adaptive IPP methods address this by reusing and updating the previous search tree rather than rebuilding it~\cite{ruckin2023informative,schmid2020efficient}. \edit{For instance, IA-TIGRIS~\cite{moon2026iatigris} combines} tree recycling with reward re-evaluation as the belief evolves. \edit{These methods provide general machinery for informative trajectory optimisation and are commonly demonstrated with uncertainty-reduction or coverage objectives. Building on IA-TIGRIS~\cite{moon2026iatigris}, we contribute a mission-conditioned reward that directs it to observations relevant to the current UGV route.}

\textit{Multi-UAV exploration and coordination} introduce a \edit{} challenge: as the team size grows, independently planned agents tend to converge on the same informative regions, leading to redundant sensing~\cite{burgard2005coordinated,zhou2023racer}. Coordination methods address this in different ways. Some rely on centralised task assignment~\cite{burgard2005coordinated}, while others use decentralised negotiation through consensus-based bundle algorithms, contract-net protocols, or multi-robot allocation schemes~\cite{chen2022consensus}. Another direction reduces overlap through spatial decomposition, e.g. by assigning agents to subregions of the environment. Zhou et al.~\cite{zhou2023racer} demonstrate that online space partitioning can support scalable decentralised exploration, while related approaches formalise Voronoi-style ownership with distributed guarantees and improved workload balancing~\cite{dong2024fast,kemna2017multi}. \edit{These works establish Voronoi-style spatial ownership as an effective way to reduce redundant multi-robot sensing. Our coordination layer uses the same established mechanism. Our contribution lies in using it to condition our proposed mission-aware reward and reduce redundant sensing of UGV-relevant regions.}

\begin{figure}[h]
\centering
\begin{tikzpicture}[
  node distance = 4mm,
  every node/.style = {font=\small},
  bluebox/.style = {
    draw = blue!50, fill = blue!7,
    rounded corners = 4pt,
    text width = 58mm,
    minimum height = 4mm,
    align = center,
    inner sep = 3pt
  },
  tealbox/.style = {
    draw = teal!60, fill = teal!7,
    rounded corners = 4pt,
    text width = 58mm,
    minimum height = 4mm,
    align = center,
    inner sep = 4pt
  },
  amberbox/.style = {
    draw = orange!60, fill = orange!8,
    rounded corners = 4pt,
    text width = 58mm,
    minimum height = 4mm,
    align = center,
    inner sep = 4pt
  },
  arr/.style = {
    -{Stealth[length=4pt]},
    thick
  },
  lbl/.style = {
    font = \scriptsize,
    inner sep = 1.5pt
  }
]
\path[use as bounding box] (-3.3cm,-4.1cm) rectangle (3.3cm,0.7cm);

\node[bluebox] (coord) {%
  \textbf{Multi-UAV coordination}\\[-1pt]
  {\scriptsize Voronoi ownership $\cdot$ Reward masking}%
};

\node[bluebox, below = of coord] (ipp) {%
  \textbf{UAV informative planning}\\[-1pt]
  {\scriptsize Mission-conditioned reward}%
};

\node[tealbox, below = of ipp] (belief) {%
  \textbf{Shared exposure belief}\\[-1pt]
  {\scriptsize Log-normal entropy-driven update}%
};

\node[amberbox, below = of belief] (ugv) {%
  \textbf{UGV planner}\\[-1pt]
  {\scriptsize Exposure-aware routing}%
};

\draw[arr]
  (coord.south) -- (ipp.north)
  node[lbl, midway, right = 2mm] {region assignment};

\draw[arr]
  ([xshift= 5mm] ipp.south) -- ([xshift= 5mm] belief.north)
  node[lbl, midway, right = 2mm] {sensor observations};

\draw[arr]
  ([xshift=-5mm] belief.north) -- ([xshift=-5mm] ipp.south)
  node[lbl, midway, left = 2mm] {reward map};

\draw[arr]
  ([xshift= 5mm] belief.south) -- ([xshift= 5mm] ugv.north)
  node[lbl, midway, right = 2mm] {exposure cost map};

\draw[arr]
  ([xshift=-5mm] ugv.north) -- ([xshift=-5mm] belief.south)
  node[lbl, midway, left = 2mm] {current UGV route};

\end{tikzpicture}
\caption{Overview of our approach. Multi-UAV coordination assigns Voronoi-based spatial ownership to each UAV planner (Sec.~\ref{sec:coordination}). Aerial observations update the shared exposure belief (Sec.~\ref{sec:belief_model}), which drives both the mission-conditioned UAV reward (Sec.~\ref{sec:ugv_ipp}) and UGV cost map (Sec.~\ref{sec:ugv_replanning}). The current UGV route feeds back to update the reward map, closing the guidance-reconnaissance loop.
}
\label{F:system}
\end{figure}

\section{Our Approach}
\label{sec:approach}

We present an adaptive planning framework for multi-UAV-assisted UGV navigation in unknown environments. Fig.~\ref{F:system} shows our approach. \edit{We consider one UGV and $N$ UAVs operating over a shared spatial domain. The robots optimise distinct but coupled objectives: the UGV plans on an eight-connected grid to minimise traversal time and cumulative exposure under the current belief (Sec.~\ref{sec:ugv_replanning}), while each UAV plans in continuous space under a motion budget to maximise a mission-conditioned reward (Sec.~\ref{sec:ugv_ipp}) serving as a proxy for improving the UGV objective.} UAV observations update a shared probabilistic exposure belief, \edit{which modifies the UGV cost map and UAV reward.} In the multi-UAV setting, we assign spatial ownership to reduce redundant sensing while preserving responsiveness to route-relevant changes.

\subsection{Shared Exposure Belief}
\label{sec:belief_model}

A shared probabilistic map is central to our framework. It provides a consistent planning interface for both the UGV and the UAV team, coupling aerial observations to ground-level routing decisions.

We consider a team of one UGV and $N$ UAVs in a bounded domain $X \subset \mathbb{R}^2$, discretised into grid $G$. Prior terrain labels for each cell $x \in G$ provide initial estimates of traversability and exposure. The latent \textit{exposure rate}  $\lambda(x) \geq 0\,[s^{-1}]$, which quantifies the instantaneous detection or hazard rate at each location, is initially unknown and must be inferred online from aerial observations. We model  $\lambda(x)$ as a log-normal random variable,
\begin{equation}
  \lambda(x) \sim
  \mathrm{LogNormal}\!\left(\mu_{\ln}(x),\,\sigma_{\ln}^2(x)\right),
  \label{E:log_normal}
\end{equation}
where $\mu_{\ln}(x)$ is initialised from terrain-class priors encoding terrain-dependent detectability, and $\sigma_{\ln}(x)$  captures epistemic uncertainty. The log-normal family is well-suited here: it yields closed-form expressions for the posterior mean $\mathbb{E}[\lambda(x)]$ and variance $\mathrm{Var}[\lambda(x)]$ used in planning, enforces strictly non-negative exposure rates, and supports multiplicative amplification by detected threat zones.

As a UAV observes a cell $x$, cumulative sensor coverage
$\mathrm{cov}(x,t) \in [0,1]$ \edit{accumulates via
\begin{equation*}
  \mathrm{cov}(x,t) = 1 - \big(1-\mathrm{cov}(x,t^-)\big)\big(1-p_{\mathrm{obs}}(x,t)\big),
\end{equation*}
where $t^-$ denotes the time of the cell's previous
update, $p_{\mathrm{obs}}(x,t) \in [0,1]$ is the instantaneous
observation confidence for cell $x$ at time $t$, from which the Bernoulli-Shannon
entropy $H(x,t) \in [0,1]$ is computed}, with $H = 1$ denoting full
uncertainty and $H = 0$ a well-observed cell. The effective log-standard-deviation is 
then updated as
\begin{equation}
  \sigma_{\ln}^{\mathrm{eff}}(x,t)
  =
  \max\!\left(
    (1 - \kappa_\sigma s(x,t))\,\sigma_{\ln}(x),\;
    \sigma_{\min}
  \right),
  \label{E:sigma_eff}
\end{equation}
where $s(x,t) = (1 - H(x,t))^{\gamma_s}$ maps entropy to observation confidence, \(\gamma_s>0\) controls the nonlinearity of this mapping, and $\kappa_\sigma \in [0,1]$ caps the  maximum uncertainty reduction per cell. The log-mean is updated analogously with a smaller coefficient $\kappa_\mu  < \kappa_\sigma$, reflecting that sensing primarily reduces uncertainty rather than shifting the estimated mean. \edit{\edit{Threat zones are revealed through deterministic proximity-based detection. A threat is instantaneously revealed once a UAV enters a predefined detection radius around its centre. Once detected, the threat zone locally amplifies the belief by a multiplicative factor $k_\tau(x,t) \geq 1$ applied to the underlying exposure distribution; before detection, $k_\tau(x,t) = 1$.}}

Both planners query the shared belief through a scalar \textit{risk-averse surrogate} that penalises high-uncertainty cells:
\begin{equation}
  \lambda_{\mathrm{risk}}(x,t)
  =
  k_\tau(x,t)\!\left(
    \mathbb{E}[\lambda(x,t)]
    + \alpha\sqrt{\mathrm{Var}[\lambda(x,t)]}
  \right),
\label{eq:lrisk}
\end{equation}
where $\alpha \geq 0$ controls risk aversion\edit{, and
$\mathbb{E}[\lambda(x,t)]$, $\mathrm{Var}[\lambda(x,t)]$ follow
in closed form from $\sigma_{\ln}^{\mathrm{eff}}(x,t)$
(Eq.~\ref{E:sigma_eff}) and the analogously updated log-mean}. Intuitively, high-uncertainty cells are treated as more hazardous than the expected value alone would suggest, creating an explicit benefit for UAV reconnaissance. Observing a cell reduces $\sigma_{\ln}^{\mathrm{eff}}$ and thus $\lambda_{\mathrm{risk}}$, lowering the UGV routing cost.

\subsection{Exposure-Aware UGV Replanning}
\label{sec:ugv_replanning}

Our goal is for the UGV to find a route that reaches the goal efficiently while minimising cumulative exposure under the current belief. We search for a path $\pi = \{x_j\}_{j=0}^K \in \Pi(x_s, x_g)$ from start $x_s$ to goal $x_g$ that minimises
\begin{equation}
  \pi_g^\star(t)
  =
  \argmin_{\pi \in \Pi(x_s, x_g)}
  \!\Big[T(\pi) + \beta\, E(\pi,t)\Big],
\label{eq:ugv_obj}
\end{equation}
where $T(\pi)$ is the total traversal time, $\beta \geq 0$ converts exposure into a time-equivalent cost, and
\begin{equation}
  E(\pi,t)
  =
  \sum_{j=0}^{K-1}
  \lambda_{\mathrm{risk}}(x_j,t)\,\Delta t_j
\label{eq:cumexp}
\end{equation}
is the cumulative exposure under the risk-averse surrogate in Eq.~\ref{eq:lrisk}. The per-step dwell time is \edit{$\Delta t_j=\frac{L_j}{v_{\max}v_{\ell(x_j)}}
$}, where $L_j \in \{\delta, \sqrt{2}\delta\}$ is the step length on the eight-connected grid\edit{, \(v_{\max}\) is the nominal UGV speed, }and \edit{$v_{\ell(x_j)}$} is a terrain-dependent speed fraction. The resulting per-cell cost thus increases both in slow terrain and in regions with high or uncertain exposure.

We solve Eq.~\ref{eq:ugv_obj} using the A*  algorithm on the cost map induced by the shared belief, with an admissible heuristic scaled by the minimum achievable per-step cost to preserve optimality. We replan when a UAV detects a new threat zone, updating both the threat overlay $k_\tau$ and the UGV route simultaneously. To maintain safe execution while replanning asynchronously, the UGV commits to a fixed-length trajectory prefix and initiates A* from its terminal cell. Each updated route reshapes the corridor used by the UAV reward map (Sec.~\ref{sec:ugv_ipp}), closing the guidance-reconnaissance loop.

\subsection{UGV-Conditioned Informative Planning}
\label{sec:ugv_ipp}

Our planner is designed to collect aerial observations that most improve the UGV's exposure-aware routing. We build on incremental adaptive IPP~\cite{moon2026iatigris}, which maintains and recycles a sampling-based search tree across planning cycles. Our contribution is a mission-conditioned reward that directs this planner toward observations relevant to the current ground route rather than to generic map coverage.

We search for a trajectory $\tau_i$ for UAV $i$ maximising the accumulated mission-conditioned reward under a motion budget $B_i$:
\begin{equation}
  \tau_i^\star
  =
  \argmax_{\tau_i}
  \sum_{x \in \mathcal{V}(\tau_i)} R(x,t)
  \quad
  \text{s.t.}\ C(\tau_i) \leq B_i,
  \label{E:uav_obj}
\end{equation}
where $\mathcal{V}(\tau_i)$ is the set of cells observed along $\tau_i$ and $R(x,t)$ is the reward defined below. 

\begin{figure}[t]
    \centering
    \includegraphics[width=1.0\columnwidth]{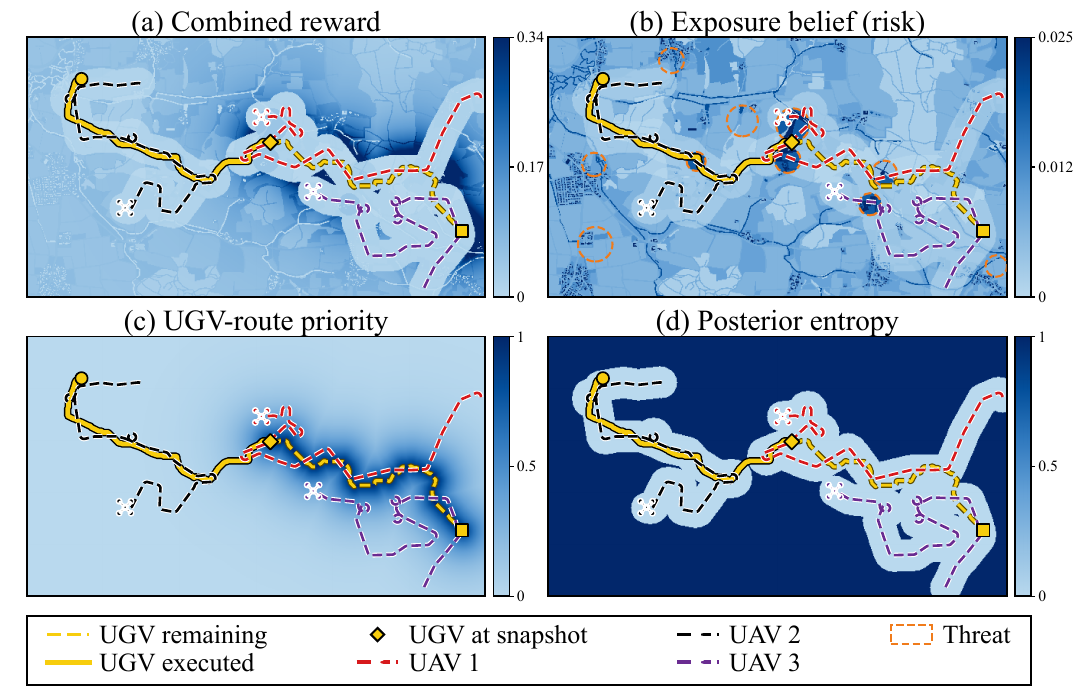}
    \caption{Component maps of the gated reward architecture at a representative mission timestep. The UGV route defines the priority map, while exposure and entropy modulate where route-relevant observations are most valuable. The resulting reward \edit{favours uncertain, exposure-relevant regions of} the UGV corridor\edit{; hence, high corridor priority alone does not imply that a UAV will directly follow the UGV path. Cells on the route that are already well observed receive low reward through the entropy term.}}
    \label{fig:gated_reward}
\end{figure}

We construct $R(x,t)$ to reflect the operational objective: corridor proximity should gate where exploration is meaningful, while exposure should modulate how urgently information is needed in those regions.  Let $P(x,t) \in [0,1]$ denote a corridor-relevance term that decays \edit{exponentially} with distance from the current UGV route $\pi_g^\star(t)$, 
\edit{\begin{equation}
  P(x,t) = \exp\!\left(-\frac{d(x,\pi_g^\star(t))}{\tau_{{P}}}\right),
  \label{E:corridor_relevance}
\end{equation}
where $d(x,\pi_g^\star(t))$ is the perpendicular
Euclidean distance from cell $x$ to the nearest point on the UGV route, and $\tau_P > 0$ controls
the rate at which corridor relevance falls off with distance.}. Let $H(x,t) \in [0,1]$ be the coverage entropy of cell $x$. \edit{The first stage combines} these into a base importance term,
\begin{equation}
  I^{\mathrm{base}}(x,t)
  =
  \left(w_p P(x,t) + w_h\right) H(x,t),
  \label{E:base_importance}
\end{equation}
\edit{where $w_p, w_h > 0$ scale the corridor-relevance term and the baseline entropy reward, respectively. This} is then passed through a priority gate,
\begin{equation}
  I(x,t)
  =
  \Big(\delta_f + (1 - \delta_f)\,P(x,t)^{\eta_p}\Big)
  \cdot I^{\mathrm{base}}(x,t),
  \label{E:priority_gate}
\end{equation}
where $\delta_f \in [0,1]$ ensures off-corridor cells retain a minimal reward, and $\eta_p > 0$ controls how sharply the gate suppresses distant regions. Intuitively, this ensures that a cell far from the UGV route receives low reward regardless of its local uncertainty or exposure level. \edit{$P(x,t)$ enters twice in distinct roles: it weights entropy linearly in
Eq.~\ref{E:base_importance} and separately gates reward by corridor proximity in Eq.~\ref{E:priority_gate},
independently of local entropy or exposure}. However, corridor proximity alone does not capture how urgently a region needs to be observed, i.e. a low-exposure corridor segment requires less immediate sensing than one near a detected threat. \edit{In a second stage, we scale} the gated importance by an exposure-dependent gain: 
\begin{equation*}
  g^{\mathrm{exp}}(x,t)
  =
  \edit{\Sig}\!\left(\edit{\kappa_g}\bigl(\bar\lambda(x,t) - \edit{\theta_\lambda}\bigr)\right),
\end{equation*}
\edit{where $\Sig(\cdot)$ denotes the standard logistic sigmoid.} A sigmoid gate $g^{\mathrm{exp}}(x,t)$ activates in high-exposure regions, where \edit{ $\bar{\lambda}(x,t)
= \min\!\left(1,\frac{\mathbb{E}[\lambda(x,t)]}{\lambda_{90}}\right)$
is the normalised posterior mean exposure, with $\lambda_{90}$ the
90th-percentile exposure value across the grid. This clipping limits
sensitivity to outliers while bounding $\bar{\lambda}(x,t)\in[0,1]$,} and $\edit{\theta_\lambda, \kappa_g}$
control the threshold and slope. The final reward is
\begin{equation}
  R(x,t)
  =
  \Big(\gamma_0 + \sigma_g\,g^{\mathrm{exp}}(x,t)\Big)
  \cdot \bar{I}(x,t)
  + w_\lambda\,\bar\lambda(x,t),
\label{eq:reward}
\end{equation}
where \edit{$\gamma_0 > 0$ is a gain applied when the exposure gate $g^{\mathrm{exp}}(x,t)$ is
inactive, $\sigma_g$ scales the exposure-dependent gain,} $\bar{I}(x,t) = I(x,t) / (w_p + w_h)$ normalises the gated importance to $[0,1]$ and $w_\lambda$ provides a small direct exposure contribution that retains visibility of isolated threat zones\edit{, since
$g^{\mathrm{exp}}$ acts multiplicatively on corridor-gated
importance while $w_\lambda\bar\lambda$ acts even where
$\bar I(x,t)\approx 0$}. As the UGV replans, $R(x,t)$ is recomputed automatically, so that UAV sensing effort realigns with the updated corridor. Fig.~\ref{fig:gated_reward} visualises how corridor priority, exposure, and entropy combine using our gated reward to concentrate UAV sensing near route-relevant, uncertain, and high-exposure regions.

\subsection{Multi-UAV Region-Conditioned Coordination}
\label{sec:coordination}

When multiple UAVs independently maximise the shared reward $R(x,t)$ \edit{in Eq.~\ref{eq:reward}}, the team converges on the same high-value corridor regions, producing redundant observations. We address this by assigning each UAV spatial ownership of a region and shaping its effective reward accordingly.

At each coordination period $t_{\mathrm{TA}}$, we partition $G$ into $N$ disjoint regions\edit{, one per UAV,} $\{\Omega_i\}$ using a Voronoi assignment on a hierarchical region graph. The graph concentrates spatial resolution near the UGV corridor, where precise ownership boundaries matter most, and uses coarser regions elsewhere to reduce assignment cost. We assign each region to the UAV with the lowest travel cost to its centroid, with a soft-claim bonus for the current owner to prevent ownership from switching back and forth when travel costs are similar. The effective reward for UAV $i$ is then
\begin{equation}
  R_i(x,t)
  =
  m_i(x,t)\,R(x,t),
  \quad
  m_i(x,t) =
  \begin{cases}
    1,   & x \in \Omega_i(t), \\
    v_f, & x \notin \Omega_i(t),
  \end{cases}
\label{eq:mask}
\end{equation}
where $v_f \in [0,1]$ is a foreign visibility parameter. Setting $v_f = 0$ yields hard coordination in which each UAV perceives no reward outside its region, maximising spatial deconfliction. Setting $v_f > 0$ retains a fraction of foreign-region reward, allowing UAVs to respond to high-value areas near their boundaries, particularly when threats emerge \edit{nearby}. Each UAV then solves \edit{the same problem as Eq.~\ref{E:uav_obj}, using the effective reward $R_i(x,t)$ in place of $R(x,t)$}:
\begin{equation*}
  \tau_i^\star
  =
  \argmax_{\tau_i}
  \sum_{x \in \mathcal{V}(\tau_i)} R_i(x,t)
  \quad
  \text{s.t.}\ C(\tau_i) \leq B_i.
\end{equation*}
Since the region assignment is recomputed each period and $R(x,t)$ is updated at each UGV replan, coordination remains aligned with the current downstream ground objective.

\begin{table}[h]
    \centering
\edit{
    \caption{Fixed method and terrain parameters. Terrain values initialise the log-normal exposure prior (Eq.~\ref{E:log_normal}) and UGV speed field from OpenStreetMap semantics.}
    \label{tab:parameters}
    \scriptsize
    \setlength{\tabcolsep}{9pt}
    \renewcommand{\arraystretch}{0.88}
    \begin{tabular}{@{}lccc@{}}
        \toprule
        Component & Parameter & \multicolumn{2}{c}{Value} \\
        \midrule
        \multicolumn{4}{c}{\textit{Method parameters}} \\
        \midrule
          Belief
          & $\gamma_s,\kappa_\sigma,\kappa_\mu$
          & \multicolumn{2}{c}{1.7, 0.95, 0.50} \\
          & $\sigma_{\min},r_{\mathrm{det}},k_\tau,\alpha$
          & \multicolumn{2}{c}{0.02, $200\,\mathrm{m}$, 6.52, 0.50} \\
          UGV
          & $\beta$
          & \multicolumn{2}{c}{$2000\,\mathrm{s}$} \\
          Reward
          & $\tau_P,w_p,w_h$
          & \multicolumn{2}{c}{$165\,\mathrm{m}$, 2.0, 0.5} \\
          & $\delta_f,\eta_p,\theta_\lambda,\kappa_g$
          & \multicolumn{2}{c}{0.10, 0.50, 0.75, 10.0} \\
          & $\gamma_0,\sigma_g,w_\lambda$
          & \multicolumn{2}{c}{0.80, 0.30, 0.005} \\
          Coordination
          & $t_{\mathrm{TA}},v_f$ (soft/hard)
          & \multicolumn{2}{c}{$30\,\mathrm{s}$, 0.4/0} \\
        \midrule
        \multicolumn{4}{c}{\textit{Terrain-dependent parameters}} \\
        \midrule
        Terrain & Log-mean offset & Log-std. & $v_\ell$ \\
        \midrule
        Road & 0.00 & 0.25 & 1.00 \\
        Grassland & 0.10 & 0.35 & 0.50 \\
        Farmland & -0.20 & 0.30 & 0.40 \\
        Wetland & 0.40 & 0.50 & 0.20 \\
        Woodland & -0.80 & 0.40 & 0.30 \\
        Forest & -1.60 & 0.45 & 0.15 \\
        Building / water & 1.20 / 0.80 & 0.40 & 0.00 \\
        Unknown & 0.20 & 0.50 & 0.60 \\
        \bottomrule
    \end{tabular}}
\end{table}

\section{Experimental Results}
\label{sec:results}

Our experiments assess the three central components of our approach. First, we evaluate whether a shared exposure belief allows UAV reconnaissance to improve downstream UGV safety beyond terrain-semantic replanning. Second, we isolate the effect of the UGV-conditioned reward and analyse whether it directs sensing toward mission-relevant regions. Third, we test whether region-conditioned coordination improves multi-UAV sensing efficiency by reducing redundant coverage while preserving responsiveness to newly relevant threat regions.

\subsection{Experimental Setup}
\label{sec:exp_setup}

\noindent \textit{Simulation setup}. We evaluate our approach in paired Monte Carlo simulation on a \(5\,\mathrm{km}\times5\,\mathrm{km}\) environment derived from OpenStreetMap data and discretised at \(10\,\mathrm{m}\) resolution. Each trial samples the UGV start and goal, the UAV initial states, and ten initially unknown threat zones, with at least three threats intersecting the initial UGV route. The UAV team collects observations that update the shared exposure belief online, while the UGV replans its route to trade traversal time against cumulative exposure.

\noindent \textit{Exposure ground truth}. Each Monte Carlo trial contains a fixed ground-truth exposure field used for evaluation. We construct this field from terrain-dependent base exposure rates and sampled threat-zone multipliers. Terrain classes derived from OpenStreetMap define both the
UGV speed fraction and the exposure prior according to
Tab.~\ref{tab:parameters}, e.g. roads allow faster travel with low base exposure, open grassland and wetland induce higher exposure, and forested areas provide lower exposure but slower traversal. Threat zones locally amplify the exposure field but are not available to planners until detected by UAV observations. Thus, the UGV is evaluated on the ground-truth field, while planners operate on the shared exposure belief described in Sec.~\ref{sec:belief_model}.

\noindent \textit{Evaluation metrics}. We evaluate UGV safety post hoc on the ground-truth exposure field using the mission objective \(O=T+\beta E\) \edit{(Eq.~\ref{eq:ugv_obj})}, realised cumulative exposure \(E\) \edit{(Eq.~\ref{eq:cumexp})}, and its base and threat-induced components \(E_{\mathrm{base}}\) and \(E_{\mathrm{threat}}\). We also report the detection probability \(P_{\mathrm{det}}(\pi)=1-\exp(-E(\pi,t))\) and tail risk \(\mathrm{CVaR}_{90}(\lambda)\)  (expected exposure rate in the worst 10\% of outcomes across trials). \edittwo{We evaluate reconnaissance quality using corridor coverage
\(C_{\mathrm{UGV}}\), the fraction of the executed UGV trajectory
eventually observed by any UAV. This is a spatial post-hoc metric which does
not require observations to occur before UGV traversal. Timing is captured
separately by \(t_{75}\), \(n_{75}\), \(T_{\mathrm{prio}}\), and
\(\mathrm{TDR}\), where \(\mathrm{TDR}\) counts priority threats detected
before the UGV reaches the corresponding route segment. We also report
redundant coverage \(C_{\mathrm{red}}\), map-level information gain
\(IG\), and useful information efficiency \(\eta_{\mathrm{useful}}\), the
non-redundant fraction of corridor- and threat-relevant \(IG\).}

\noindent \textit{Implementation}. \edit{Tab.~\ref{tab:parameters} shows the method parameters used in all experiments, based on a separate tuning set before evaluation.} Unless stated otherwise, each result reports \(100\) Monte Carlo paired trials under identical random seeds, sensing budgets, vehicle models, and planner parameters. UAVs fly at \(30\,\mathrm{m/s}\) with Dubins constraints and a \(200\,\mathrm{m}\) sensing footprint with \(N=3\) UAVs as default.
The UGV replans when a threat zone is detected, triggering a full UAV reward map recomputation. Corridor relevance \(P(x,t)\) \edit{(Eq.~\ref{E:corridor_relevance})} decays with distance from the current route and is combined with entropy and exposure using fixed reward parameters across the paired trials. For multi-UAV experiments, we recompute region ownership every \(t_{\mathrm{TA}}=30\,\mathrm{s}\).

\noindent \edit{\textit{Baselines}. We compare against two single-UAV baselines. \textit{Scouting path-aware}~\cite{rockenbauer2025traversing}  \edittwo{selects unobserved targets on an optimistic UGV path computed from the partially observed cost map while the UGV executes and replans. It is path-aware but not a pure path-following scout, as the optimistic path can differ from the executed route, and the UAV reaches targets subject to Dubins constraints.}
\textit{Coverage$\rightarrow$plan} first surveys the entire environment using a Dubins boustrophedon path before UGV execution.}


\subsection{Shared Exposure Belief}
\label{sec:exp_exposure}

This experiment evaluates whether coupling UAV reconnaissance to a shared exposure belief improves downstream UGV safety beyond terrain-semantic replanning alone. We compare four progressively informed conditions. The \textit{Static} baseline uses no UAV support. \textit{Terrain-semantic} uses UAV reconnaissance and UGV replanning from semantic traversability and high-cost threat avoidance, but does not model the exposure distribution. \textit{Exposure-aware} is our full shared-belief method. \textit{Oracle} is an optimistic reference with full prior knowledge of the threat-augmented exposure field.

  \begin{figure}[t]
  \centering
  \resizebox{0.85\columnwidth}{!}{%
      \begin{tikzpicture}
  \tikzset{every node/.style={font=\footnotesize}}
  \begin{axis}[
    width=7.35cm,
    height=3.6cm,
    scale only axis,
    xmin=0, xmax=400,
    ymin=0, ymax=5.2,
    xlabel={Time [s]},
    ylabel={Cumulative exposure \(E\)},
    xtick={0,100,200,300,400},
    ytick={0,1,2,3,4,5},
    grid=major,
    grid style={draw=oiGray!18},
    axis line style={draw=oiGray!70},
    tick style={draw=oiGray!70},
    tick label style={font=\scriptsize},
    label style={font=\footnotesize},
    legend columns=1,
    legend style={
    at={(0.03,0.97)},
    anchor=north west,
    draw=cGray!70,
    fill=white,
    fill opacity=0.85,
    text opacity=1,
    rounded corners=1pt,
    inner sep=2pt,
    font=\scriptsize,
    },
  ]
  \addplot[name path=staticlo, draw=none, forget plot] coordinates {
    (0,0.00) (20,0.16) (40,0.27) (60,0.50) (80,0.77)
    (100,1.08) (120,1.43) (140,1.85) (160,2.28) (180,2.69)
    (200,3.07) (220,3.39) (240,3.70) (260,3.95) (280,4.16)
    (300,4.31) (320,4.38) (340,4.40) (360,4.40)
  };
  \addplot[name path=statichi, draw=none, forget plot] coordinates {
    (0,0.00) (20,0.17) (40,0.32) (60,0.70) (80,1.04)
    (100,1.36) (120,1.76) (140,2.28) (160,2.69) (180,3.14)
    (200,3.51) (220,3.84) (240,4.15) (260,4.42) (280,4.65)
    (300,4.83) (320,4.91) (340,4.93) (360,4.93)
  };
  \addplot[oiVermilion!12, forget plot] fill between[of=staticlo and statichi];

  \addplot[name path=terrainlo, draw=none, forget plot] coordinates {
    (0,0.00) (20,0.10) (40,0.20) (60,0.31) (80,0.43)
    (100,0.57) (120,0.70) (140,0.84) (160,0.97) (180,1.10)
    (200,1.24) (220,1.37) (240,1.48) (260,1.58) (280,1.66)
    (300,1.72) (320,1.75) (340,1.75) (360,1.76)
  };
  \addplot[name path=terrainhi, draw=none, forget plot] coordinates {
    (0,0.00) (20,0.16) (40,0.29) (60,0.39) (80,0.52)
    (100,0.65) (120,0.79) (140,0.93) (160,1.06) (180,1.20)
    (200,1.34) (220,1.47) (240,1.59) (260,1.69) (280,1.78)
    (300,1.85) (320,1.89) (340,1.91) (360,1.92)
  };
  \addplot[oiGreen!12, forget plot] fill between[of=terrainlo and terrainhi];

  \addplot[name path=exposurelo, draw=none, forget plot] coordinates {
    (0,0.00) (20,0.09) (40,0.17) (60,0.23) (80,0.30)
    (100,0.38) (120,0.45) (140,0.53) (160,0.63) (180,0.71)
    (200,0.78) (220,0.85) (240,0.92) (260,0.99) (280,1.06)
    (300,1.14) (320,1.20) (340,1.25) (360,1.28) (380,1.30)
    (400,1.30)
  };
  \addplot[name path=exposurehi, draw=none, forget plot] coordinates {
    (0,0.00) (20,0.14) (40,0.22) (60,0.28) (80,0.35)
    (100,0.44) (120,0.52) (140,0.60) (160,0.70) (180,0.79)
    (200,0.86) (220,0.93) (240,1.00) (260,1.06) (280,1.14)
    (300,1.23) (320,1.30) (340,1.35) (360,1.38) (380,1.41)
    (400,1.41)
  };
  \addplot[oiBlue!12, forget plot] fill between[of=exposurelo and exposurehi];

  \addplot[oiGray, thick, dashed] coordinates {
    (0,0.00) (20,0.10) (40,0.17) (60,0.22) (80,0.28)
    (100,0.35) (120,0.41) (140,0.47) (160,0.55) (180,0.60)
    (200,0.67) (220,0.71) (240,0.76) (260,0.82) (280,0.87)
    (300,0.94) (320,1.00) (340,1.05) (360,1.05)
  };
  \addlegendentry{Oracle}

  \addplot[oiVermilion, thick, dashed] coordinates {
    (0,0.00) (20,0.15) (40,0.29) (60,0.60) (80,0.91)
    (100,1.22) (120,1.59) (140,2.07) (160,2.50) (180,2.91)
    (200,3.30) (220,3.62) (240,3.93) (260,4.19) (280,4.42)
    (300,4.57) (320,4.65) (340,4.67) (360,4.67)
  };
  \addlegendentry{Static}

  \addplot[oiGreen, thick] coordinates {
    (0,0.00) (20,0.13) (40,0.24) (60,0.35) (80,0.47)
    (100,0.61) (120,0.74) (140,0.88) (160,1.02) (180,1.15)
    (200,1.29) (220,1.42) (240,1.53) (260,1.63) (280,1.72)
    (300,1.79) (320,1.82) (340,1.83) (360,1.84)
  };
  \addlegendentry{Terrain-semantic}

  \addplot[oiBlue, thick] coordinates {
    (0,0.00) (20,0.12) (40,0.19) (60,0.25) (80,0.32)
    (100,0.41) (120,0.48) (140,0.57) (160,0.67) (180,0.75)
    (200,0.82) (220,0.89) (240,0.96) (260,1.03) (280,1.10)
    (300,1.18) (320,1.25) (340,1.30) (360,1.34) (380,1.35)
    (400,1.35)
  };
  \addlegendentry{Exposure-aware}

  \end{axis}
  \end{tikzpicture}
  }
  \caption{Cumulative UGV exposure over mission time. Solid and dashed curves show means over $100$ paired trials; shaded regions denote \(95\%\) confidence intervals. Our exposure-aware replanning scheme consistently reduces accumulated exposure under the same aerial sensing budget.}
  \label{fig:cumulative_exposure}
  \end{figure}
  
\begin{table}[t]
  \centering
  \caption{Oracle-normalised safety performance. Using our shared exposure belief closes substantially more of the oracle gap.}
  \label{tab:exposure_model}
  \resizebox{\columnwidth}{!}{%
  \begin{tabular}{@{}lccc@{}}
    \toprule
    Condition & \(O_{\mathrm{oracle}}\) [\%] \(\uparrow\) &
    \(E_{\mathrm{gt,oracle}}\) [\%] \(\uparrow\) &
    \(P_{\mathrm{det,oracle}}\) [\%] \(\uparrow\) \\
    \midrule
    Static & 15.2 & 12.1 & 41.7 \\
    Terrain-semantic & 52.8 & 45.4 & 60.7 \\
    Exposure-aware & \textbf{77.0} & \textbf{72.0} & \textbf{79.8} \\
    Oracle & 100.0 & 100.0 & 100.0 \\
    \bottomrule
  \end{tabular}}
\end{table}

Fig.~\ref{fig:cumulative_exposure} shows the cumulative UGV exposure over mission time. UAV-supported replanning substantially reduces exposure relative to the static baseline, showing the benefit of updating the UGV route from aerial observations. The exposure-aware method further lowers total exposure, indicating that explicitly modelling exposure using our approach enables reconnaissance to improve UGV route safety.

\begin{figure*}[!t]
    \centering
    \includegraphics[width=0.98\linewidth]{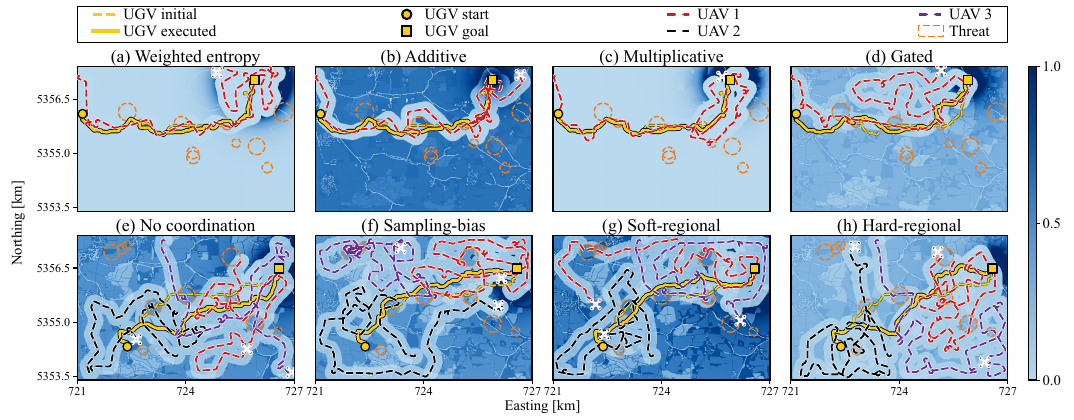}
    \vspace*{-2.5mm}
    \caption{Example final mission states for the reward \edit{and} coordination studies. \edit{Panels in the same row use the same paired trial.} The background shows the final UAV reward map (darker blue indicates higher reward), curves show UGV and UAV trajectories, and orange circles indicate threat zones. Top row compares reward architectures (Sec.~\ref{sec:exp_reward}): unlike generic fusion, our proposed gated reward preserves corridor-focused sensing while reallocating attention to exposure-relevant regions, improving UGV safety. Bottom row compares UAV coordination strategies (Sec.~\ref{sec:exp_coordination}): increasing coordination strength reduces overlap and redundant sensing, while soft visibility retains limited responsiveness to high-value regions outside a UAV's assigned domain.}
    \label{fig:reward_qualitative}
\end{figure*}

 \begin{table*}[t]
    \centering
    \caption{\edittwo{Reward ablations and planning baselines for one
    UAV over \(100\) trials. Mission costs are means \\ and \(t_{75}\) is the mean \(\pm95\%\) confidence interval. Bold marks the best result within reward ablations.}}
    \label{tab:reward_arch}
    \scriptsize
    \renewcommand{\arraystretch}{0.9}
    \resizebox{0.92\linewidth}{!}{%
    \begin{tabular}{@{}lrrrrrrr@{}}
      \toprule
      Method & \(O\) [s] \(\downarrow\) &
      \(\beta E_{\mathrm{base}}\) [s] \(\downarrow\) &
      \(\beta E_{\mathrm{threat}}\) [s] \(\downarrow\) &
      \(C_{\mathrm{UGV}}\) [\%] \(\uparrow\) &
      \(t_{75}\) [s] \(\downarrow\) &
      \(n_{75}\) [\%] \(\uparrow\) &
      \(\mathrm{TDR}\) [\%] \(\uparrow\) \\
      \midrule
      \multicolumn{8}{l}{\edittwo{\textit{Reward architecture ablations}}}\\
      Weighted entropy
        & 2536.80 & 1707.80 & \textbf{495.04}
        & \textbf{72.3} & 242.8$\pm$14.3
        & \textbf{74} & \textbf{41} \\
      Additive
        & 2775.49 & 1746.28 & 695.60
        & 64.4 & 248.1$\pm$19.3
        & 43 & 33 \\
      Multiplicative
        & 2565.74 & \textbf{1612.28} & 623.14
        & 69.4 & 254.0$\pm$22.1
        & 44 & 34 \\
      Gated
        & \textbf{2496.50} & 1631.71 & 531.52
        & 71.6 & \textbf{239.0$\pm$17.0}
        & 64 & \textbf{41} \\
       \midrule
       \multicolumn{8}{l}{\edittwo{\textit{External planning baselines}}} \\
      \edit{Scouting path-aware~\cite{rockenbauer2025traversing}}
        & \edit{2858.7} & \edit{1800.9} & \edit{740.3}
        & \edit{43.2} & \edit{217.3$\pm$26.4}
        & \edit{24} & \edit{36.4} \\
      \edit{Coverage$\rightarrow$plan}
        & \edit{1355.8} & \edit{1028.3} & \edit{3.1}
        & \edit{100.0} & \edit{1726.9$\pm$124.1}
        & \edit{100} & \edit{100.0} \\
      \bottomrule
    \end{tabular}}
  \end{table*}

Tab.~\ref{tab:exposure_model} shows \edit{safety performance} relative to the full-information oracle. \edit{The key comparison is between \textit{Terrain-semantic} and \textit{Exposure-aware}, which use the same sensing budget and replanning capability. Our shared exposure belief improves oracle-normalised performance, reducing the mission objective by 31\% and cumulative exposure by 38\%. This is driven mainly by a 55\% reduction in threat-related exposure, compared with 33\% in base exposure, while $\mathrm{CVaR}_{90}$ also decreases substantially. These results show that the shared exposure belief translates aerial observations into route-relevant risk reduction beyond semantic map updates.}

\subsection{UGV-Conditioned Reward}
\label{sec:exp_reward}

We next evaluate the UAV reward architecture \edit{with two additional baselines} to test whether UGV-conditioned sensing improves mission-level navigation. Here, we consider $N = 1$ UAV to isolate reward mapping from coordination effects. All methods use the same belief layers and differ only in how entropy, corridor relevance, and exposure are fused into a sensing reward. We set hyperparameters for each method using the same manual tuning procedure and number of tuning trials, so that our gated reward's larger parameter count does not constitute an unfair advantage. \textit{Weighted entropy} scales uncertainty by a corridor-priority overlay and represents a corridor-focused sensing policy. \textit{Additive} and \textit{Multiplicative} combine the same terms through generic fusion rules. \textit{Gated} is our UGV-conditioned reward in Eq.~\ref{eq:reward}.

Tab.~\ref{tab:reward_arch} shows that generic reward fusion does not automatically produce mission-relevant reconnaissance. \textit{Additive} spreads sensing effort too broadly, leading to the highest mission cost and lowest corridor coverage. \textit{Multiplicative} improves base exposure \edit{relative to \textit{Weighted entropy}, but sacrifices corridor coverage and performs worse on threat-related exposure, corridor-refinement timing, and priority-threat detection}. In contrast, our \textit{Gated} reward achieves the lowest mission objective while maintaining nearly the same corridor coverage and priority-threat detection rate as \textit{Weighted entropy}. \edit{\textit{Weighted entropy} achieves lower threat-related exposure, but \textit{Gated} provides the best downstream UGV objective overall.} It also reaches the \(t_{75}\) corridor-refinement milestone earliest.
\edit{Compared with the existing path-aware UAV-UGV method of Rockenbauer et al.~\cite{rockenbauer2025traversing}, \textit{Gated} achieves lower mission cost and higher corridor coverage, while \textit{Coverage$\rightarrow$plan} attains the lowest exposure only after a \(3228.9\,\mathrm{s}\) pre-scouting phase. These baselines highlight the trade-off offered by online mission-conditioned sensing relative to path-only scouting and exhaustive pre-coverage.}


\begin{table*}[t]
  \centering
  \caption{Coordination method comparison for $3$ UAVs over $100$ trials. Region-conditioned masking sharply reduces redundant coverage and improves information efficiency, while soft masking preserves the priority-threat detection rate.}
  \label{tab:coordination} 
  \scriptsize
  \renewcommand{\arraystretch}{0.86}
  \resizebox{0.92\linewidth}{!}{%
  \begin{tabular}{@{}lrrrrrrr@{}}
    \toprule
    Regime & IG [\%] \(\uparrow\) & \(\eta_{\mathrm{useful}}\) [\%] \(\uparrow\) &
    \(C_{\mathrm{red}}\) [\%] \(\downarrow\) & \(C_{\mathrm{UGV}}\) [\%] \(\uparrow\) &
    \(O\) [s] \(\downarrow\) & \(T_{\mathrm{prio}}\) [s] \(\downarrow\) &
    \(\mathrm{TDR}\) [\%] \(\uparrow\) \\
    \midrule
    No coordination & 41.02 & 66.2 & 38.8 & 85.3 & 1845.10 & 145.02 & 50 \\
    Sampling-bias & 42.56 & 71.8 & 29.2 & 85.1 & 1876.19 & 130.43 & 51 \\
    Soft-regional & 46.91 & 92.9 & 5.0 & 89.1 & 1816.81 & 128.40 & \textbf{55} \\
    Hard-regional & \textbf{47.39} & \textbf{94.9} & \textbf{3.7} & \textbf{90.6} &
    \textbf{1804.91} & \textbf{127.64} & 45 \\
    \bottomrule
  \end{tabular}}
\end{table*}

The top row of Fig.~\ref{fig:reward_qualitative} visualises representative trajectories for the reward formulations. \textit{Additive} diffuses sensing away from the UGV, while \textit{Gated} preserves corridor support and targets exposure-relevant areas, consistent with the findings in Tab.~\ref{tab:reward_arch}. Together, these results support our claim that conditioning the UAV objective on the current UGV route focuses sensing on mission-relevant information.

\subsection{Region-Conditioned Multi-UAV Coordination}
\label{sec:exp_coordination}

Finally, we validate \edit{region-conditioned coordination in
the multi-UAV setting, examining how increasing coordination
strength trades reduced redundancy against responsiveness to
newly relevant threat regions as the team scales.} We compare four regimes: \textit{No coordination}, where all UAVs optimise the same global reward; \textit{Sampling-bias}, where Voronoi ownership biases sampling but reward visibility remains global; \textit{Soft-regional}, where each UAV sees a reduced reward fraction outside its region \edit{via $v_f$ in Eq.~\ref{eq:mask}}; and \textit{Hard-regional}, where foreign-region reward is fully masked. 


Tab.~\ref{tab:coordination} shows that sampling-bias only partly \edit{reduces} redundant sensing: \edit{reward visibility remains global.} Soft- and hard-regional \edit{masking} reduce redundant coverage from \(38.8\%\) to \(5.0\%\) and \(3.7\%\), respectively, while increasing useful information efficiency from \(66.2\%\) to \edit{over} \(92\%\). \edit{Hard-regional coordination also achieves the lowest mean mission objective.} The bottom row of Fig.~\ref{fig:reward_qualitative} \edit{illustrates the resulting trajectory separation. Namely, without coordination, UAVs revisit the same high-reward corridor regions, whereas regional masking distributes them across complementary areas.}

The results expose the \edit{redundancy-responsiveness trade-off}. \textit{Hard-regional} gives the best spatial efficiency and mission objective, but lowers priority-threat detection rates. \textit{Soft-regional} retains \edit{cross-region visibility, yielding highest \(\mathrm{TDR}\) while keeping redundancy low. Regional ownership thus prevents convergence on the same reward peaks, while soft visibility preserves responsiveness near region boundaries.}

\begin{figure}
\definecolor{figAxis}{RGB}{107,114,128}
\definecolor{cNoCoord}{RGB}{68,119,170}
\definecolor{cBias}{RGB}{34,136,51}
\definecolor{cSoft}{RGB}{204,102,119}

\resizebox{0.985\linewidth}{!}{%
\begin{tikzpicture}
\tikzset{every node/.style={font=\footnotesize}}

\path[use as bounding box] (-0.52cm,-0.70cm) rectangle (8.32cm,5.4cm);

\def\panelW{3.85cm}
\def\panelH{2.18cm}
\def\panelXGap{4.48cm}
\def\panelYGap{2.78cm}
\def\boxOffA{-0.24}
\def\boxOffB{-0.08}
\def\boxOffC{0.08}
\def\boxOffD{0.24}
\def\boxW{0.115}

\pgfplotsset{
  scaleplot/.style={
    width=\panelW,
    height=\panelH,
    scale only axis,
    boxplot/draw direction=y,
    xmin=0.55, xmax=4.45,
    xtick={1,2,3,4},
    xticklabels={1,2,4,6},
    xlabel={},
    ymajorgrids,
    grid style={draw=figAxis!18},
    axis line style={draw=figAxis!80, line width=0.35pt},
    tick style={draw=figAxis!80, line width=0.3pt},
    tick align=inside,
    major tick length=1.0pt,
    xtick pos=bottom,
    ytick pos=left,
    tick label style={font=\scriptsize},
    label style={font=\footnotesize},
    xlabel style={font=\footnotesize, yshift=0.4ex},
    ylabel style={
      font=\footnotesize,
      at={(axis description cs:-0.105,0.5)},
      anchor=south
    },
    every boxplot/.style={solid, line width=0.42pt},
  }
}

\newcommand{\ScaleBox}[8]{%
  \addplot+[
    boxplot prepared={
      draw position=#2,
      lower whisker=#3,
      lower quartile=#4,
      median=#5,
      upper quartile=#6,
      upper whisker=#7,
      box extend=\boxW
    },
    draw=#1!90!black,
    fill=#1!58,
    mark=none
  ] coordinates {};
}

\newcommand{\LegendEntry}[3]{%
  \draw[#1!90!black, fill=#1!58, line width=0.38pt]
    (#2,0) rectangle ++(0.16,0.095);
  \node[anchor=west,font=\scriptsize] at ({#2+0.22},0.047) {#3};
}

\begin{scope}[shift={(0.05cm,5.22cm)}]
  \draw[
    draw=black!55,
    fill=white,
    rounded corners=0pt,
    line width=0.25pt
  ] (-0.60,-0.16) rectangle (8.3,0.22);

  \LegendEntry{cNoCoord}{-0.5}{No coordination}
  \LegendEntry{cBias}{2.0}{Sampling bias}
  \LegendEntry{cSoft}{4.20}{Soft regional}
  \LegendEntry{cHard}{6.20}{Hard regional}
\end{scope}

\begin{axis}[
  scaleplot,
  at={(0, \panelYGap)},
  anchor=south west,
  ymin=10, ymax=76,
  ytick={10,30,50,70},
]
\ScaleBox{cNoCoord}{1}{14.0}{17.8}{19.4}{21.9}{27.5}{}

\ScaleBox{cNoCoord}{2+\boxOffA}{22.0}{29.1}{31.7}{33.6}{39.4}{}
\ScaleBox{cBias}{2+\boxOffB}{25.7}{30.3}{33.6}{35.8}{41.2}{}
\ScaleBox{cSoft}{2+\boxOffC}{27.9}{31.8}{34.1}{35.9}{42.2}{(0,43.8)}
\ScaleBox{cHard}{2+\boxOffD}{25.7}{31.7}{34.3}{35.8}{42.2}{}

\ScaleBox{cNoCoord}{3+\boxOffA}{33.2}{39.6}{42.2}{45.0}{53.1}{}
\ScaleBox{cBias}{3+\boxOffB}{36.8}{43.5}{45.5}{48.1}{55.2}{}
\ScaleBox{cSoft}{3+\boxOffC}{39.6}{47.2}{49.1}{51.0}{55.1}{(0,40.4)}
\ScaleBox{cHard}{3+\boxOffD}{40.8}{46.6}{49.4}{51.3}{55.5}{}

\ScaleBox{cNoCoord}{4+\boxOffA}{30.6}{39.3}{42.2}{46.9}{56.3}{}
\ScaleBox{cBias}{4+\boxOffB}{35.2}{45.8}{48.4}{51.7}{61.4}{(0,32.2) (0,36.0)}
\ScaleBox{cSoft}{4+\boxOffC}{45.8}{55.8}{62.0}{65.2}{74.0}{}
\ScaleBox{cHard}{4+\boxOffD}{49.4}{57.8}{61.9}{65.0}{72.5}{}
\end{axis}

\begin{axis}[
  scaleplot,
  at={(\panelXGap, \panelYGap)},
  anchor=south west,
  ymin=-15, ymax=285,
  ytick={0,100,200},
]
\ScaleBox{cNoCoord}{1}{0.0}{0.0}{0.0}{0.0}{0.0}{}

\ScaleBox{cNoCoord}{2+\boxOffA}{0.0}{15.9}{21.2}{29.8}{53.5}{}
\ScaleBox{cBias}{2+\boxOffB}{0.0}{5.4}{11.9}{20.5}{39.0}{}
\ScaleBox{cSoft}{2+\boxOffC}{0.0}{0.0}{0.7}{3.4}{19.2}{(0,10.5) (0,13.2)}
\ScaleBox{cHard}{2+\boxOffD}{0.0}{0.7}{1.4}{4.0}{23.2}{(0,12.0) (0,17.0)}

\ScaleBox{cNoCoord}{3+\boxOffA}{25.0}{52.2}{64.1}{71.3}{106.9}{}
\ScaleBox{cBias}{3+\boxOffB}{8.7}{29.8}{39.6}{50.9}{73.9}{}
\ScaleBox{cSoft}{3+\boxOffC}{0.1}{7.3}{10.6}{13.3}{29.1}{}
\ScaleBox{cHard}{3+\boxOffD}{0.1}{7.5}{8.5}{10.6}{33.7}{}

\ScaleBox{cNoCoord}{4+\boxOffA}{56.1}{123.4}{161.0}{196.6}{275.0}{}
\ScaleBox{cBias}{4+\boxOffB}{26.5}{83.8}{116.8}{141.2}{203.9}{}
\ScaleBox{cSoft}{4+\boxOffC}{5.0}{33.1}{39.6}{47.6}{70.0}{}
\ScaleBox{cHard}{4+\boxOffD}{5.0}{27.8}{35.7}{38.3}{62.0}{}
\end{axis}

\begin{axis}[
  scaleplot,
  at={(0,0)},
  anchor=south west,
  xlabel={Number of UAVs},
  ymin=40, ymax=410,
  ytick={100,200,300,400},
]
\ScaleBox{cNoCoord}{1}{151}{218}{260}{282}{360}{(0,390)}

\ScaleBox{cNoCoord}{2+\boxOffA}{73}{137}{193}{275}{340}{}
\ScaleBox{cBias}{2+\boxOffB}{73}{131}{182}{251}{360}{}
\ScaleBox{cSoft}{2+\boxOffC}{95}{142}{183}{229}{350}{}
\ScaleBox{cHard}{2+\boxOffD}{92}{167}{206}{250}{372}{}

\ScaleBox{cNoCoord}{3+\boxOffA}{55}{86}{105}{141}{228}{(0,238) (0,260) (0,302) (0,331) (0,340)}
\ScaleBox{cBias}{3+\boxOffB}{55}{98}{120}{146}{207}{(0,219) (0,246) (0,282) (0,326) (0,349)}
\ScaleBox{cSoft}{3+\boxOffC}{61}{102}{122}{146}{216}{(0,225) (0,247) (0,285) (0,323)}
\ScaleBox{cHard}{3+\boxOffD}{56}{102}{123}{150}{217}{(0,238) (0,266) (0,282)}

\ScaleBox{cNoCoord}{4+\boxOffA}{61}{103}{170}{227}{320}{}
\ScaleBox{cBias}{4+\boxOffB}{57}{104}{161}{225}{307}{}
\ScaleBox{cSoft}{4+\boxOffC}{56}{98}{137}{213}{316}{}
\ScaleBox{cHard}{4+\boxOffD}{57}{99}{146}{218}{300}{}
\end{axis}

\begin{axis}[
  scaleplot,
  at={(\panelXGap,0)},
  anchor=south west,
  xlabel={Number of UAVs},
  ymin=-15, ymax=370,
  ytick={0,100,200,300},
]
\ScaleBox{cNoCoord}{1}{104}{206}{250}{287}{334}{}

\ScaleBox{cNoCoord}{2+\boxOffA}{25}{82}{198}{246}{344}{}
\ScaleBox{cBias}{2+\boxOffB}{25}{92}{159}{266}{335}{}
\ScaleBox{cSoft}{2+\boxOffC}{23}{96}{124}{170}{247}{(0,300) (0,306)}
\ScaleBox{cHard}{2+\boxOffD}{25}{93}{146}{183}{269}{}

\ScaleBox{cNoCoord}{3+\boxOffA}{9}{70}{95}{155}{284}{(0,315)}
\ScaleBox{cBias}{3+\boxOffB}{10}{70}{102}{156}{268}{(0,303) (0,315)}
\ScaleBox{cSoft}{3+\boxOffC}{9}{63}{98}{150}{276}{(0,313) (0,352)}
\ScaleBox{cHard}{3+\boxOffD}{2}{81}{104}{135}{226}{(0,250) (0,285) (0,312)}

\ScaleBox{cNoCoord}{4+\boxOffA}{23}{73}{128}{199}{270}{}
\ScaleBox{cBias}{4+\boxOffB}{20}{74}{151}{211}{260}{}
\ScaleBox{cSoft}{4+\boxOffC}{22}{62}{126}{215}{320}{}
\ScaleBox{cHard}{4+\boxOffD}{23}{85}{136}{189}{277}{}
\end{axis}

\node[
  anchor=north west,
  font=\footnotesize\bfseries,
  fill=white,
  fill opacity=0.88,
  text opacity=1,
  inner sep=1pt
] at (0.08cm,4.86cm) {(a) [\%]};

\node[
  anchor=north west,
  font=\footnotesize\bfseries,
  fill=white,
  fill opacity=0.88,
  text opacity=1,
  inner sep=1pt
] at (4.56cm,4.86cm) {(b) [\%]};

\node[
  anchor=north west,
  font=\footnotesize\bfseries,
  fill=white,
  fill opacity=0.88,
  text opacity=1,
  inner sep=1pt
] at (0.08cm,2.08cm) {(c) [s]};

\node[
  anchor=north west,
  font=\footnotesize\bfseries,
  fill=white,
  fill opacity=0.88,
  text opacity=1,
  inner sep=1pt
] at (4.56cm,2.08cm) {(d) [s]};

\end{tikzpicture}
    }
    \vspace*{-1.5mm}
    \caption{Scalability study with UAV team size over $100$ trials. We compare (a) information gain, (b) redundant area coverage, (c) corridor refinement time \(t_{75}\), and (d) priority-threat detection time \(T_{\mathrm{prio}}\). More UAVs collect more information, but without coordination they increasingly observe the same high-reward areas. Region-conditioned masking reduces this redundancy and preserves parallel sensing.}
    \label{fig:results_scalability}
\end{figure}

\edit{We further evaluate coordination for $N\in\{1,2,4,6\}$ UAVs using the same paired Monte Carlo protocol. Fig.~\ref{fig:results_scalability} shows diminishing information-gain returns with team size, while redundant coverage grows rapidly without region-conditioned masking. Soft and hard masking largely suppress this redundancy and preserve useful parallel sensing.}

The timing metrics show that more UAVs do not automatically improve all mission-relevant outcomes. Corridor refinement and priority-threat detection \edit{improve initially but then saturate}. Priority-threat detection time worsens beyond four UAVs under \textit{Hard-regional}, suggesting that \edit{Voronoi fragmentation can isolate UAVs from threats in neighbouring regions and increase response latency}. This reinforces the trade-off \edit{between hard masking for spatial efficiency and soft visibility for responsiveness near region boundaries.}

\noindent\textit{Limitations}. \edittwo{Our evaluation assumes idealised sensing and communication. Namely, we use an entropy-driven belief update, noise-free UAV observations, and instantaneous lossless belief sharing. The reward is route-conditioned but not lead-time-aware, so route cells may receive reward even if the UGV arrives before observation. Further tests with sensing noise, missed detections, packet loss, and communication delay showed degradation in absolute performance but preserved the qualitative trends between methods. These results provide additional evidence that the observed trends are not specific to the nominal idealised setting, although deployment validation and lead-time-aware sensing rewards remain future work.}

\section{Conclusion}
\label{sec:conclusion}

In this paper, we presented a multi-UAV informative planning approach for exposure-aware UGV navigation in unknown threat-augmented environments. Our approach maintains a shared exposure belief that couples aerial observations to ground-level route planning. The UGV uses this belief to replan safer routes online, while the UAVs prioritise \edit{sensing} relevant to the current ground route. To scale the approach to multiple UAVs, we use region-conditioned coordination that assigns spatial ownership and reduces redundant sensing.

\edit{Results show} that coupling aerial reconnaissance to a shared exposure belief improves UGV safety compared to terrain-semantic replanning. Our proposed UGV-conditioned reward directs sensing toward mission-relevant regions, and  region-conditioned coordination substantially reduces redundant coverage while maintaining responsiveness to threat regions. Future work will \edit{extend controlled robustness tests to deployment-level sensing and communication conditions and validate the approach on heterogeneous air-ground hardware.}



\bibliographystyle{IEEEtran}

\bibliography{glorified,new}

\end{document}